\documentclass{article}
\usepackage{ijcai23}

\usepackage{times}
\usepackage{soul}
\usepackage{url}
\usepackage[hidelinks]{hyperref}
\usepackage[utf8]{inputenc}
\usepackage[small]{caption}
\usepackage{graphicx}
\usepackage{amsmath}
\usepackage{amsthm}
\usepackage{booktabs}
\usepackage{algorithm}
\usepackage{algorithmic}
\usepackage[switch]{lineno}
\usepackage{bm}
\usepackage{bbm}
\usepackage{subfigure}

\usepackage{xcolor}

\linenumbers

\title{Federated Learning for Short Text Clustering}

\newcommand{\modelname}{\textbf{FSTC}}
\author{
Mengling Hu, Chaochao Chen, Weiming Liu, Xinting Liao, and Xiaolin Zheng, \\
Zhejiang University, China\\
\texttt{\{humengling,zjuccc,21831010,xintingliao,xlzheng\}@zju.edu.cn}
}

\begin{document}

\maketitle

\setlength{\floatsep}{4pt plus 4pt minus 1pt}
\setlength{\textfloatsep}{4pt plus 2pt minus 2pt}
\setlength{\intextsep}{4pt plus 2pt minus 2pt}
\setlength{\dbltextfloatsep}{3pt plus 2pt minus 1pt}
\setlength{\dblfloatsep}{3pt plus 2pt minus 1pt}
\setlength{\abovecaptionskip}{3pt}
\setlength{\belowcaptionskip}{2pt}
\setlength{\abovedisplayskip}{2pt plus 1pt minus 1pt}
\setlength{\belowdisplayskip}{2pt plus 1pt minus 1pt}

\begin{abstract}
Short text clustering has been popularly studied for its significance in mining valuable insights from many short texts.
In this paper, we focus on the federated short text clustering (FSTC) problem, i.e., clustering short texts that are distributed in different clients, which is a realistic problem under privacy requirements.
Compared with the centralized short text clustering problem that short texts are stored on a central server, the FSTC problem has not been explored yet.
To fill this gap, we propose a Federated Robust Short Text Clustering (\modelname) framework.
\modelname~includes two main modules, i.e., \textit{robust short text clustering module} and \textit{federated cluster center aggregation module}.
The robust short text clustering module aims to train an effective short text clustering model with local data in each client.
We innovatively combine optimal transport to generate pseudo-labels with Gaussian-uniform mixture model to ensure the reliability of the pseudo-supervised data.
The federated cluster center aggregation module aims to exchange knowledge across clients without sharing local raw data in an efficient way.
The server aggregates the local cluster centers from different clients and then sends the global centers back to all clients in each communication round.
Our empirical studies on three short text clustering datasets demonstrate that \modelname~significantly outperforms the federated short text clustering baselines.
\end{abstract}

\section{Introduction}
Short text clustering has been proven to be beneficial in many applications, such as, news recommendation \cite{wu2022personalized}, opinion mining \cite{stieglitz2018social}, stance detection \cite{li2022unsupervised}, etc. 
Existing short text clustering models
\cite{xu2017self,hadifar2019self,rakib2020enhancement,zhang2021supporting} all assume that the short texts to be clustered are stored on a central server where their models are trained.
However, the assumption may be invalid when the data is distributed among many clients and gathering the data on a central server is not feasible due to privacy regulations or communication concerns
\cite{magdziarczyk2019right,DBLP:conf/aistats/McMahanMRHA17,ffcm2022morris}.
For example, a multi-national company, with several local markets, sells similar commodities in all markets.
Each local market has text data about their customers, e.g., personal information, purchased items, reviews, etc.
As text clustering is one of the most fundamental tasks in text mining, the company wishes to cluster text data from all markets for text mining, which can mine more reliably valuable insights compared with only clustering local data.
The mined valuable information can further guide the marketing strategy of the company.
Because of strict privacy regulations, the company is not allowed to gather all data in a central server, e.g., European customer data is not allowed to be transferred to most countries outside of Europe \cite{otto2018regulation,ffcm2022morris}.

In this paper, we focus on the federated short text clustering (FSTC) problem.
Federated learning is widely used to enable collaborative learning across a variety of clients without sharing local raw data \cite{tan2022fedproto}.
Federated clustering is a kind of federated learning setting whose goal is to cluster data that is distributed among multiple clients.
Unlike popular federated supervised classification task, federated unsupervised clustering task is less explored and existing federated clustering methods cannot work well with short texts.
Specifically, existing federated clustering methods can be divided into two types, i.e., the k-means based federated clustering methods
\cite{kumar2020federated,pedrycz2021federated,dennis2021heterogeneity,ffcm2022morris}
and the deep neural network based federated clustering method \cite{chung2022federated}.
The former methods are not applicable to short texts because short texts often have very sparse representations that lack expressive ability.
It is beneficial to utilize deep neural network to enrich the short text representations for better clustering performance \cite{xu2017self,hadifar2019self,zhang2021supporting}.
However, the latter deep learning based method \cite{chung2022federated} cannot cope with real-world noisy data well.
Therefore, existing federated clustering methods cannot be utilized to solve the FSTC problem.

The FSTC problem has not been explored yet, possibly because short texts are sparse and noisy, which makes it difficult to cluster short texts in the federated environment.
\cite{mcmahan2017communication} proposes FedAvg to substitute synchronized stochastic gradient descent for the federated learning of deep neural networks.
Combining the state-of-the-art short text clustering models with FedAvg \cite{mcmahan2017communication} seems to be a reasonable way to solve the FSTC problem.
However, the combination cannot solve the FSTC problem well.
Firstly, existing short text clustering models
\cite{xu2017self,hadifar2019self,rakib2020enhancement,zhang2021supporting} cannot learn sufficiently discriminative representations due to lacking supervision information, causing limited clustering performance.
Secondly, FedAvg needs to aggregate models in every communication round, causing limited communication efficiency.
In summary, there are two main challenges, i.e., \textbf{CH1}: How to provide supervision information for discriminative representation learning, and promote better clustering performance?
\textbf{CH2}: How to exchange knowledge across clients in a more efficient way?

To address the aforementioned challenges, in this paper, we propose \modelname, a novel framework for federated short text clustering.
In order to provide supervision information (solving \textbf{CH1}) and exchange knowledge (solving \textbf{CH2}), we utilize two modules in \modelname, i.e., \textit{robust short text clustering module} and \textit{federated cluster center aggregation module}.
The robust short text clustering module aims to tackle the first challenge by generating pseudo-labels as the supervision information.
We leverage optimal transport to generate pseudo-labels, and introduce Gaussian-uniform mixture model to estimate the probability of correct labeling for more reliable pseudo-supervised data.
The federated cluster center aggregation module aims to tackle the second challenge by aggregating cluster centers rather than models in every communication round.
We use the locally generated pseudo-labels to divide the clusters of a client for obtaining the local cluster centers, and align the local centers of all clients for collaboration.

We summarize our main contributions as follows:
(1) We propose a novel framework, i.e., \modelname, for federated short text clustering.
To our best knowledge, we are the first to address short text clustering problem in the federated learning setting.
(2) We propose an end-to-end model for local short text clustering, which can learn more discriminative representations with reliable pseudo-supervised data and promote better clustering performance.
(4) We conduct extensive experiments on three short text clustering datasets and the results demonstrate the superiority of \modelname.
\section{Related Work}
\subsection{Short Text Clustering}
Short text clustering is not a trivial task due to the weak signal contained in each text instance.
The existing short text clustering methods can be divided into tree kinds: (1) traditional methods, (2) deep learning methods, and (3) deep joint clustering methods.
The traditional methods \cite{scott1998text,salton1983introduction} often obtain very sparse representations that lack discriminations.
The deep learning method \cite{xu2017self} leverages pre-trained word embeddings \cite{mikolov2013distributed} and deep neural network to enrich the representations.
However, it does not combine a clustering objective with the deep representation learning, which prevents the learned representations from being appropriate for clustering.
The deep joint clustering methods \cite{hadifar2019self,zhang2021supporting} integrate clustering with deep representation learning to learn the representations that are appropriate for clustering.
Moreover, \cite{zhang2021supporting} utilizes the pre-trained SBERT \cite{reimers2019sentence} and contrastive learning to learn discriminative representations.
However, the learned representations are still insufficiently discriminative due to the lack of supervision information \cite{hu2021learning}.
As a contrast, in this work, we combine pseudo-label technology with Gaussian-uniform mixture model to provide reliable supervision to learn more discriminative representations.

\subsection{Federated Clustering}
Federated clustering aims to cluster data that is distributed among multiple clients.
Unlike the popularity of federated supervised classification task, federated unsupervised clustering is underdeveloped.
Existing federated clustering methods can be divided into two types, i.e., the k-means based federated clustering methods \cite{kumar2020federated,pedrycz2021federated,dennis2021heterogeneity,ffcm2022morris} and the deep neural network based federated clustering method \cite{chung2022federated}.
\cite{kumar2020federated} extends k-means algorithm to the federated setting, they propose calculating a weighted average of local cluster centers to update the global cluster centers, the weights are given by the samples number in clusters.
\cite{pedrycz2021federated} introduces a fuzzy c-means federated clustering, which uses fuzzy assignments as weights instead of the samples number in clusters.
\cite{dennis2021heterogeneity} introduces a one-shot k-means federated clustering method which utilize k-means to aggregate and update the global cluster centers.
\cite{ffcm2022morris} proposes a federated fuzzy c-means clustering method which is similar to \cite{kumar2020federated,pedrycz2021federated,dennis2021heterogeneity}.
The deep neural network based federated clustering methods are not well studied.
Only \cite{chung2022federated} develops a new generative model based clustering method in the federated setting, based on IFCA \cite{ghosh2020efficient} algorithm.
However, \cite{chung2022federated} shows that the method can obtain good clustering performance for synthetic datasets, but always fails when training with real-world noisy data.
As a contrast, in this work, our method can obtain good clustering performance for the real-world noisy data. 
\section{Methodology}
\begin{figure*}[t]
  \centering
  \includegraphics[width=1\linewidth]{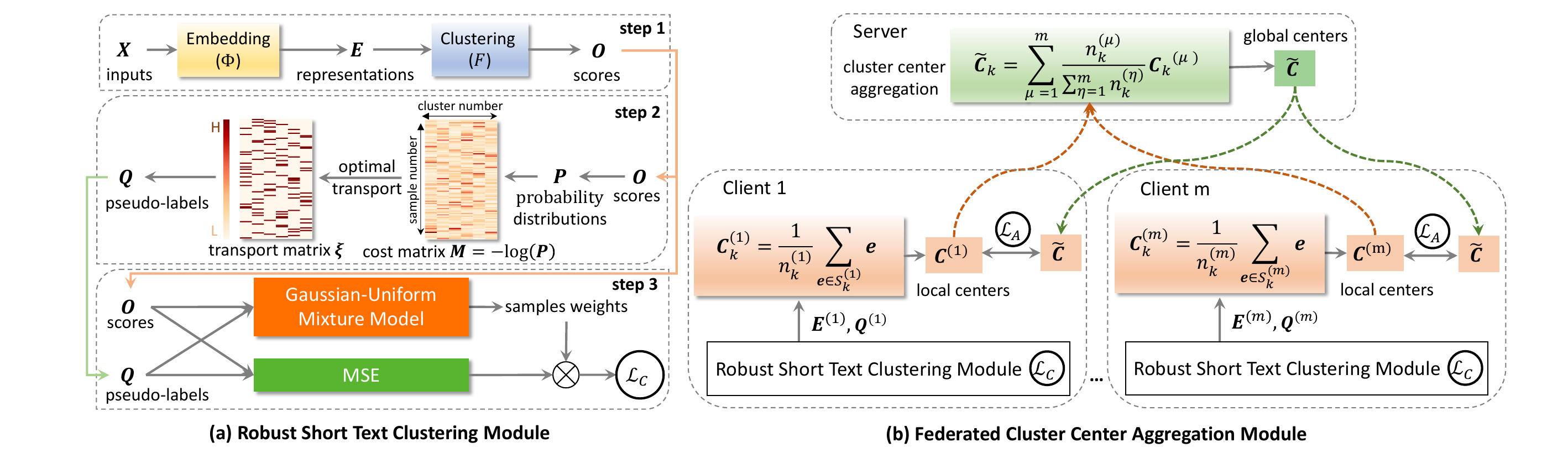}
  \vspace{-0.5cm}
  \caption{{An overview of \modelname.}
  }
  \vspace{-0.2cm}
  \label{fig:model}
\end{figure*}

\subsection{An Overview of \modelname}
The goal of \modelname~is to collaboratively train a global short text clustering model with the raw data stored locally in multiple clients.
The overall structure of our proposed \modelname~is illustrated in Fig.\ref{fig:model}.
\modelname~consists of two main modules, i.e., \textit{robust short text clustering module}  and \textit{federated cluster center aggregation module}.
The robust short text clustering module aims to train a short text clustering model with local data in a client.
The federated cluster center aggregation module aims to efficiently exchange information across clients without sharing their local raw data.
In the end, we can obtain the global model by averaging the final local models.
We will introduce these two modules in detail later.

\subsection{Robust Short Text Clustering Module}
We first introduce the robust short text clustering module.
Although the deep joint clustering methods \cite{hadifar2019self,zhang2021supporting} on short text clustering are popular these days, their clustering performance is still limited.
The reason is that lacking supervision information prevents those methods from learning more discriminative representations \cite{hu2021learning}.
Therefore, to provide supervision information for short text clustering, we propose to generate \textit{pseudo-labels} for guiding the local model training, that is, unsupervised training samples turn into pseudo-supervised training samples.
Moreover, because the pseudo-labels are inevitably noisy, we design a robust learning objective for fully exploiting the pseudo-supervised training samples.
An overview of the robust short text clustering module is shown in Fig.\ref{fig:model}(a), which mainly has three steps: \textbf{Step 1}: predicting cluster assignment scores, \textbf{Step 2}: generating pseudo-labels, and \textbf{Step 3}: obtaining a robust objective.
Note that, we divide the robust short text clustering module into three steps just for the sake of introduction convenience, the whole module is trained in an end-to-end way.
We will introduce the details below.

\paragraph{Step 1: predicting cluster assignment scores.}
This step aims to predict and provide cluster assignment scores of the inputs for the other two steps.
For inputs $\bm{X}$, we adopt SBERT \cite{reimers2019sentence} as the embedding network $\Phi$ to obtain the representations, i.e., $\Phi{(\bm{X})}=\bm{E}\in \mathbbm{R}^{N\times D}$, where $N$ denotes batch size and $D$ is the dimension of the representations.
We utilize fully connected layers as the clustering network $F$ to predict cluster assignment scores, i.e., $F(\bm{E})=\bm{O}\in \mathbbm{R}^{N\times K}$, where $K$ is the number of clusters.

\paragraph{Step 2: generating pseudo-labels.}
This step aims to generate pseudo-labels of all samples by excavating information from the cluster assignment scores and provide the pseudo-labels for \textbf{Step 3}.
To begin with, we use softmax \cite{bridle1990probabilistic} to normalize the scores $\bm{O}$ for obtaining the cluster assignment probability distributions $\bm{P}\in \mathbbm{R}^{N\times K}$.
For each sample, we expect its generated pseudo-label distribution to align its predicted probability distribution.
Specifically, we denote the pseudo-labels as $\bm{Q}\in \mathbbm{R}^{N\times K}$.
We adopt KL-divergence minimization for aligning the pseudo-label distributions and the predicted probability distributions.
Meanwhile, to avoid the trivial solution that all samples are assigned to one cluster, we add the constraint that the label assignments partition data equally \cite{asano2019self,caron2020unsupervised}.
Formally, the objective is as follows:
\begin{equation}
\begin{aligned}
\min_{\bm{Q}} {\rm KL} (\bm{Q} \parallel \bm{P}) \quad s.t.\,\, \sum_{i=1}^N{\bm{Q}_{ij}}&=\frac{N}{K}, \nonumber
\end{aligned}
\end{equation}
where
\begin{equation}
\begin{aligned}
\label{equ:kl}
    {\rm KL} (\bm{Q} \parallel \bm{P})=-\sum_{i=1}^N{\sum_{j=1}^K{\bm{Q}_{ij}\log{\bm{P}_{ij}}}} + \sum_{i=1}^N{\sum_{j=1}^K{\bm{Q}_{ij}\log{\bm{Q}_{ij}}}}.
\end{aligned}
\end{equation}
We turn this objective into a discrete optimal transport problem \cite{cuturi2013sinkhorn}, i.e.,
\begin{equation}
    \begin{aligned}
        \bm{\xi}^* = \underset{\bm{\xi}}{\mathrm{argmin}}\, {\langle \bm{\xi}, \bm{M} \rangle + \epsilon H(\bm{\xi})},\\
        s.t., \,\, \bm{\xi}\mathbbm{1}_K=\bm{a},\bm{\xi}^T\mathbbm{1}_N=\bm{b}, \bm{\xi}\geq 0,
    \end{aligned}
\end{equation}
where $\bm{\xi}=\bm{Q}$ denotes the transport matrix, $\bm{M}=-\log\bm{P}$ denotes the cost matrix,
$H(\bm{\xi})=\sum_{i=1}^N{\sum_{j=1}^K}\bm{\xi}_{ij}\log{\bm{\xi}_{ij}}$ denotes the entropy constraint,
$\langle.\rangle$ is the Frobenius dot-product between two matrices,
$\epsilon$ is the balance hyper parameter,
 $\bm{a}=\mathbbm{1}_N$, $\bm{b}=\frac{N}{K}\mathbbm{1}_K$. 
We apply the Sinkhorn algorithm \cite{cuturi2013sinkhorn} to solve the optimal transport problem.
Specifically, we introduce Lagrangian multipliers, then the objective turns into:
\begin{equation}
\begin{aligned}
    \min_{\bm{\xi}}
    \langle \bm{\xi},\bm{M} \rangle
    +\epsilon H(\bm{\xi})
    -\bm{f}^T(\bm{\xi}\mathbbm{1}_K
    -\bm{a}) -\bm{g}^T(\bm{\xi}^T\mathbbm{1}_N -\bm{b}),
\label{equ:ot_loss}
\end{aligned}
\end{equation}
where $\bm{f}$ and $\bm{g}$ are both Lagrangian multipliers.
Taking the differentiation of Equation (\ref{equ:ot_loss}) on the variable $\bm{\xi}$, we can obtain:
\begin{equation}
\begin{aligned}
    \bm{\xi}=\text{diag}(\bm{u})\bm{\mathcal{K}}\text{diag}(\bm{v}),
\label{equ:ot_matrix}
\end{aligned}
\end{equation}
where $\bm{u}=\exp{({\bm{f}}/{\epsilon}-{1}/{2})}$,
$\bm{\mathcal{K}}=\exp{(-{\bm{M}}/{\epsilon})}$,
and $\bm{v}=\exp{({\bm{g}}/{\epsilon}-{1}/{2})}$.
Taking Equation (\ref{equ:ot_matrix}) back to the original constraints $\bm{\xi}\mathbbm{1}_K=\bm{a},\bm{\xi}^T\mathbbm{1}_N=\bm{b}$, we can obtain:
\begin{equation}
\begin{aligned}
    \bm{u}=\bm{a}\oslash \bm{\mathcal{K}}\bm{v},
\label{equ:u}
\end{aligned}
\end{equation}
\begin{equation}
\begin{aligned}
    \bm{v}=\bm{b}\oslash \bm{\mathcal{K}}^T\bm{u},
\label{equ:v}
\end{aligned}
\end{equation}
where $\oslash$ is the Hadamard division.
Through iteratively solving Equation (\ref{equ:u}) and Equation (\ref{equ:v}), we can obtain the transport matrix $\bm{\xi}$ on Equation (\ref{equ:ot_matrix}).
Furthermore, although we let $\bm{\xi}=\bm{Q}$ before, we square the values in $\bm{\xi}$ for obtaining more reliable pseudo-labels $\bm{Q}$ \cite{xie2016unsupervised}.
Specifically, $\bm{Q}$ is formulated as:
\begin{equation}
\label{eq:pseudo_labels}
    \begin{aligned}
        {Q}_{ik}=\frac{{\xi}_{ik}^2}{\sum_{k^{'}=1}^K \xi_{ik^{'}}^2}.
    \end{aligned}
\end{equation}
We show the optimization scheme of pseudo-label generation in Algorithm~\ref{al:sinkhorn}.
\begin{algorithm}[t]
    \caption{Pseudo-label Generation.}
    \label{al:sinkhorn}
    \textbf{Input}: The cluster assignment scores $\bm{O}$; the balance hyper-parameter $\epsilon$; batch size $N$; cluster number $K$.\\
    \textbf{Procedure}:
    \begin{algorithmic}[1]
        \STATE{Calculate the cluster assignment probability distributions $\bm{P}=\text{softmax}(\bm{O})$}
        \STATE{Calculate the cost matrix $\bm{M}=-\log{\bm{P}}$.}
        \STATE{Calculate $\bm{\mathcal{K}}=\exp{(-{\bm{M}}/{\epsilon})}$.}
        \STATE{Randomly initialize vectors $\bm{u}\in \mathbbm{R}^N$ and $\bm{v}\in \mathbbm{R}^K$.}
        \FOR{$i=1$ to $t$}
            \STATE{Update $\bm{u}$ by Equation (\ref{equ:u}).}
            \STATE{Update $\bm{v}$ by Equation (\ref{equ:v}).}
        \ENDFOR
        \STATE{Calculate $\bm{\xi}$ by Equation (\ref{equ:ot_matrix}).}
        \STATE{Calculate $\bm{Q}$ by Equation (\ref{eq:pseudo_labels}).}
        \RETURN{$\bm{Q}$}
    \end{algorithmic}
\end{algorithm}

\paragraph{Step 3: obtaining a robust objective.}
This step aims to design a robust objective to fully exploit the pseudo-supervised samples.
Although the generated pseudo-labels can be helpful to learn more discriminative representations, not all of the pseudo-labels are correct and the wrong pseudo-labels may prevent our model from achieving better performance.
Thus, to mitigate the influence of wrong pseudo-labels, we propose to estimate the probability of correct labeling, and use the probability to weight corresponding pseudo-supervised sample.
Specifically, inspired by \cite{deepgum}, we use a Gaussian-uniform mixture model to model the distribution of a pseudo-label $\bm{Q}_i$ conditioned by its cluster assignment score $\bm{O}_i$:
\begin{equation}
    \begin{aligned}
        p(\bm{Q}_i|\bm{O}_i)=\pi \mathcal{N}(\bm{Q}_i;\bm{O}_i,\bm{\Sigma})+(1-\pi)\mathcal{U}(\bm{Q}_i;\gamma),
    \end{aligned}
\end{equation}
where $\mathcal{N}(.)$ denotes a multivariate Gaussian distribution and $\mathcal{U}(.)$ denotes a uniform distribution.
The Gaussian distribution models the correct pseudo-labels while the uniform distribution models the wrong pseudo-labels.
$\pi$ is the prior probability of a correct pseudo-label, $\bm{\Sigma}\in \mathbbm{R}^{K\times K}$ is the the covariance matrix of the Gaussian distribution, and $\gamma$ is the normalization parameter of the uniform distribution.
The posterior probability of correct labeling for $i$-th sample is,
\begin{equation}
\label{eq:update_r}
    \begin{aligned}
        r_i\leftarrow \frac{\pi \mathcal{N}(\bm{Q}_i;\bm{O}_i,\bm{\Sigma})}{\pi \mathcal{N}(\bm{Q}_i;\bm{O}_i,\bm{\Sigma})+(1-\pi)\gamma}.
    \end{aligned}
\end{equation}
The parameters of Gaussian-uniform mixture models are $\bm{\theta}=\{\pi, \bm{\Sigma}, \gamma\}$.
We update these parameters with:
\begin{equation}
\label{eq:update_sigma}
    \begin{aligned}
        \bm{\Sigma}\leftarrow \sum_{i=1}^N r_i\bm{\delta}_i\bm{\delta}_i^T,
    \end{aligned}
\end{equation}
\begin{equation}
\label{eq:update_pi}
    \begin{aligned}
        \pi\leftarrow \sum_{i=1}^N r_i/N,
    \end{aligned}
\end{equation}
\begin{equation}
\label{eq:update_gamma}
    \begin{aligned}
        \frac{1}{\gamma}\leftarrow \prod_{k=1}^K 2\sqrt{3(C_{2k}-C_{1k}^2)},
    \end{aligned}
\end{equation}
where $\bm{\delta}_i=\bm{Q}_i-\bm{O}_i$,
$C_{1k}\leftarrow \frac{1}{N}\sum_{i=1}^N \frac{1-r_i}{1-\pi}\delta_{ik}$, and
$C_{2k}\leftarrow \frac{1}{N}\sum_{i=1}^N\frac{1-r_i}{1-\pi}\delta_{ik}^2$.
For more details about Gaussian-uniform mixture model, please refer to \cite{coretto2016robust,deepgum}.
For further mitigating the influence of wrong pseudo-labels, we discard samples with probability of correct labeling less than 0.5 following \cite{gu2020spherical}, i.e., the weight of a pseudo-supervised sample is defined as,
\begin{equation}
\label{eq:update_w}
    \begin{aligned}
    w_i=\begin{cases}
        r_i,&\text{if } r_i\geq 0.5,\\
        0,&\text{otherwise}.
        \end{cases}
    \end{aligned}
\end{equation}
With generated pseudo-labels and samples weights, our robust clustering objective is defined as,
\begin{equation}
\label{eq:lc}
    \begin{aligned}
        \mathcal{L}_C = \frac{1}{\sum_{i=1}^N w_i}\sum_{i=1}^N w_i\parallel \bm{Q}_i-\bm{O}_i \parallel_2^2.
    \end{aligned}
\end{equation}
We adopt mean square error (MSE) as the objective to train our model with pseudo-supervised data, because MSE is more robust to label noise than the cross entropy loss in classification task \cite{ghosh2017robust}.
Through the three steps, we can learn more discriminative representations and achieve better short text clustering performance with local data.

\subsection{Federated Cluster Center Aggregation Module}
We then introduce the federated cluster center aggregation module.
Existing short text clustering methods \cite{xu2017self,hadifar2019self,rakib2020enhancement,zhang2021supporting} all assume full access to data, i.e., the data is stored on a central server.
However, the data may be distributed among many clients (e.g., companies), and gathering the data to a central server is not always feasible due to the privacy or communication concerns \cite{DBLP:conf/aistats/McMahanMRHA17}.
Therefore, to enable collaborative learning across a variety of clients without sharing local raw data, we propose the federated cluster center aggregation module.
To ensure the communication efficiency, our federated learning module communicates the cluster centers rather than the model parameters during training process, inspired by \cite{tan2022fedproto}.
However, the partition of clusters is unknown in a clustering scenario, causing unavailable cluster centers.
Thus, we use the locally generated pseudo-labels to divide the clusters of a client.
An overview of the federated learning module is shown in Fig.\ref{fig:model}(b).
The server receives local centers from all clients and then averages these centers for obtaining global centers.
The clients receive global centers and update their local centers by minimizing the clustering loss and the distance between global centers and local centers.
We will provide the details below.

We assume that there are $m$ clients, each client has $K$ clusters.
The sample $i$ will be grouped into $k$-th cluster if the $k$-th entry of its pseudo-label $\bm{Q}_{i}$ is the largest.
We denote the samples belonging to $k$-th cluster as $S_k$.
We obtain the local cluster centers by averaging the representations of samples in every cluster set.
For client $\mu$, the center of cluster $k$ is computed as follows,
\begin{equation}
\label{eq:local_center}
    \begin{aligned}
        \bm{C}^{(\mu)}_{k}=\frac{1}{n_k^{(\mu)}}\sum_{\bm{e}\in S_k^{(\mu)}}{\bm{e}},
    \end{aligned}
\end{equation}
where $n_k^{(\mu)}$ is the number of samples assigned to the $k$-th cluster of client $\mu$, i.e., $n_k^{(\mu)}=|S_k^{(\mu)}|$.
We obtain the global cluster centers by weighted averaging the local centers of all clients.
The weights are given by the number of samples assigned to the local clusters.
For cluster $k$, the global center is computed as follows,
\begin{equation}
\label{eq:global_center}
    \begin{aligned}
    \widetilde{\bm{C}}_k=\sum_{\mu=1}^m{\frac{n_k^{(\mu)}}{\sum_{\eta=1}^mn_k^{(\eta)}}\bm{C}_k^{(\mu)}}.
    \end{aligned}
\end{equation}
We expect the local centers $\bm{C}^{(1)},\bm{C}^{(2)},...,\bm{C}^{(m)}$ to approach global centers $\widetilde{\bm{C}}$ to align the local centers of all clients for exchanging information across clients.
We achieve this aim by the alignment loss, for client $\mu$, the alignment loss is defined as follows,
\begin{equation}
\label{eq:loss_a}
    \begin{aligned}
        \mathcal{L}_A=\Vert \bm{C}^{(\mu)}-\widetilde{\bm{C}}\Vert^2.
    \end{aligned}
\end{equation}
\section{Putting Together}
The total loss of a client could be obtained by combining the clustering loss and the alignment loss.
That is, the loss of a client is given as:
\begin{equation}
\label{eq:loss_model}
    \begin{aligned}
        \mathcal{L}=\mathcal{L}_C + \lambda\mathcal{L}_A,
    \end{aligned}
\end{equation}
where $\lambda$ is a hyper-parameter to balance the two losses.
In the end, we obtain the global model by averaging the final client models.
The cluster assignments are the column index of the largest entry in each row of predicted scores $\bm{O}$.
By doing this, \modelname~achieves effective and efficient short text clustering with the raw data stored locally in multiple clients.
We show the optimization scheme of \modelname~in Algorithm~\ref{al:model}.
\begin{algorithm}[t]
\caption{The optimization scheme of \modelname.}
\label{al:model}
\textbf{Input}: The inputs of $m$ clients: $\bm{X}^{(\mu)}$ for $\mu=1,...,m$.\\
\textbf{ServerUpdate}:
\begin{algorithmic}[1]
\STATE{Initialize client models with the same parameters.}
\STATE{Aggregate representations from $m$ clients and perform k-means on them to initialize $\bm{Q}^{(\mu)}$ for $\mu=1,...,m$.}
\STATE{Initialize $\bm{C}^{(\mu)}$ by Equation (\ref{eq:local_center}) for $\mu=1,...,m$.}
\STATE{Initialize $\widetilde{\bm{C}}$ by Equation (\ref{eq:global_center}).}
\STATE{Initialize $\bm{r}^{(\mu)}=\mathbbm{1}_N$ for $\mu=1,...,m$.}
\FOR{each round}
    \FOR{each client $\mu$}
        \STATE{$\bm{C}^{(\mu)}$ $\leftarrow$ ClientUpdate$(\mu,\widetilde{\bm{C}})$ by Algorithm \ref{al:clientupdate}.}
    \ENDFOR
    \STATE{Update $\widetilde{\bm{C}}$ by Equation (\ref{eq:global_center}).}
\ENDFOR
\STATE{Aggregate the parameters of $m$ client models to obtain the global model,
i.e., $\Phi=\sum_{\mu=1}^m\alpha^{(\mu)}\Phi^{(\mu)}$ and 
$F=\sum_{\mu=1}^m\alpha^{(\mu)}F^{(\mu)}$, where $\alpha^{(\mu)}$ is the samples proportion of client $\mu$ in all clients.}
\RETURN{$\Phi$ and $F$.
}
\end{algorithmic}
\end{algorithm}

\begin{algorithm}[t]
\caption{ClientUpdate$(\mu,\widetilde{\bm{C}})$}
\label{al:clientupdate}
\begin{algorithmic}[1]
    \FOR{each local iteration}
        \STATE{Update $\bm{Q}^{(\mu)}$ by Algorithm~\ref{al:sinkhorn} throughout the whole training iterations in a logarithmic distribution \cite{asano2019self}.}
        \FOR{$j=1$ to $\tau$}
            \STATE{Update $\bm{r}^{(\mu)}$ by Equation (\ref{eq:update_r}).}
            \STATE{Update $\bm{\theta}^{(\mu)}$ by Equation (\ref{eq:update_sigma}),(\ref{eq:update_pi}),(\ref{eq:update_gamma}).}
        \ENDFOR
        \STATE{Update $\bm{w}^{(\mu)}$ by Equation (\ref{eq:update_w}).}
        \STATE{Update $\Phi^{(\mu)}$ and $F^{(\mu)}$ by minimizing Equation (\ref{eq:loss_model}).}
        \STATE{Update $\bm{C}^{(\mu)}$ by Equation (\ref{eq:local_center}).}
    \ENDFOR
    \RETURN{$\bm{C}^{(\mu)}$}
\end{algorithmic}
\end{algorithm}

\section{Experiment}
In this section, we conduct experiments on three real-world datasets to answer the following questions:
(1) \textbf{RQ1}: How does our approach perform compared with the federated short text clustering baselines?
(2) \textbf{RQ2}: How do the Gaussian-uniform mixture model and the federated cluster center aggregation module contribute to the performance improvement?
(3) \textbf{RQ3}: How does the performance of \modelname~vary with different values of the hyper-parameters?
(4) \textbf{RQ4}: How dose the performance of \modelname~vary with different numbers of clients?
(5) \textbf{RQ5}: How dose the performance of \modelname~vary with  different non-IID levels?
\subsection{Datasets}
We conduct extensive experiments on three popularly used real-world datasets.
The details of each dataset are as follows.
\textbf{AgNews} \cite{rakib2020enhancement} is a subset of AG's news corpus collected by \cite{zhang2015character} which consists of 8,000 news titles in 4 topic categories.
\textbf{StackOverflow} \cite{xu2017self} consists of  20,000 question titles associated with 20 different tags, which is randomly selected from the challenge data published in Kaggle.com\footnote{https://www.kaggle.com/c/predict-closed-questions-on-stack-overflow/}.
\textbf{Biomedical} \cite{xu2017self} is composed of 20,000 paper titles from 20 different topics and it is selected from the challenge data published in BioASQ's official website\footnote{http://participants-area.bioasq.org/}.
The detailed statistics of these datasets are shown in Table \ref{ta:dataset}.

We consider the experiments of both IID partition and non-IID partition in multiple clients.
\textbf{IID Partition}: The client number is set to $m=\{2,4,8,10\}$, the data is shuffled and evenly partitioned into multiple clients.
\textbf{Non-IID Partition}: The client number is set to $m=2$, the data is shuffled and partitioned into 2 clients with different proportions \{6:4, 7:3, 8:2, 9:1\} which correspond to different non-IID levels $\rho=\{1, 2, 3, 4\}$.

\begin{table}
\centering
	\begin{tabular}{c | c c c c c}
	\hline
		Dataset & \#clusters & \#samples & \#words \\
		\hline
		AgNews & 4 & 8,000 & 23\\
		StackOverflow & 20 & 20,000 & 8\\
		Biomedical & 20 & 20,000 & 13\\
		\hline
	\end{tabular}
	\caption{The statistics of the datasets. \#words denotes the average word number per sample.}
	\label{ta:dataset}
\end{table}

\subsection{Evalutation Metrics}
We report two widely used performance metrics of text clustering, i.e., accuracy (ACC) and normalized mutual information (NMI), following former short text clustering literatures \cite{xu2017self,hadifar2019self,zhang2021supporting}.
Accuracy is defined as:
\begin{equation}
    \begin{aligned}
        ACC=\frac{\sum_{i=1}^N{\mathbbm{1}_{y_i=map(\hat{y}_i)}}}{N},
    \end{aligned}
\end{equation}
where $y_i$ and $\hat{y}_i$ are the ground truth label and the predicted label for a given text $x_i$ respectively, $map()$ maps each predicted label to the corresponding target label by Hungarian algorithm\cite{papadimitriou1998combinatorial}.
Normalized mutual information is defined as:
\begin{equation}
    \begin{aligned}
        NMI(Y, \hat{Y})=\frac{I(Y, \hat{Y})}{\sqrt{H(Y)H(\hat{Y})}},
    \end{aligned}
\end{equation}
where $Y$ and $\hat{Y}$ are the ground truth labels and the predicted labels respectively, $I()$ is the mutual information and $H()$ is the entropy.

\subsection{Experiment Settings}
We build our framework with PyTorch \cite{pytorch} and train the local models using the Adam optimizer \cite{kingma2014adam}.
We choose distilbert-base-nli-stsb-mean-tokens in Sentence Transformer library \cite{reimers2019sentence} to embed the short texts, and the maximum input length is set to $32$.
The learning rate is set to $5\times10^{-6}$ for optimizing the embedding network, and $5\times10^{-4}$ for optimizing the clustering network.
The dimensions of the text representations is set to $D=768$.
The batch size is set to $N=200$.
The hyper-parameter $\epsilon$ is set to $0.1$.
We study the effect of hyper-parameter $\lambda$ by varying it in $\{0.001, 0.01, 0.1, 1, 10\}$.
The communication rounds is set to $40$ and the local iterations is set to $100$.
Following previous short text clustering researches \cite{xu2017self,hadifar2019self,rakib2020enhancement,zhang2021supporting}, we set the clustering numbers to the ground-truth category numbers.
Moreover, we adopt the same augmentation strategy with \cite{zhang2021supporting} for achieving better representation learning.

\subsection{Baselines}
We compare our proposed approach with the following federated short text clustering baselines.
\textbf{FBOW}: We apply k-FED \cite{dennis2021heterogeneity} on the BOW \cite{scott1998text} representations.
\textbf{FTF-IDF}: We apply k-FED \cite{dennis2021heterogeneity} on the TF-IDF \cite{salton1983introduction} representations.
\textbf{FSBERT}: We apply k-means on the representations embedded by SBERT \cite{reimers2019sentence}.
Note that, as the BOW representations and TF-IDF representations reveal the raw texts, they cannot be transmitted to the central server and directly applying k-means.
While SBERT representations do not reveal the raw texts, they can be transmitted to the central server and directly applying k-means.
\textbf{FSCCL}: We combine FedAvg \cite{mcmahan2017communication} with SCCL
\cite{zhang2021supporting}.
SCCL is one of the state-of-the-art short text clustering models, it utilizes SBERT \cite{reimers2019sentence} as the backbone, introduces instance-wise contrastive learning to support clustering, and uses the clustering objective proposed in \cite{xie2016unsupervised} for deep joint clustering.
We use its released code\footnote{https://github.com/amazon-science/sccl} for achieving the local model.

\subsection{Federated Clustering Performance (RQ1)}
\begin{table*}
\centering
  \footnotesize
  \begin{tabular}{ccccccc}
    \toprule
     &\multicolumn{2}{c}{\textbf{AgNews}} &\multicolumn{2}{c}{\textbf{Stackoverflow}} &\multicolumn{2}{c}{\textbf{Biomedical}}\\
     \cmidrule{2-7}
     & \text{ACC} & \text{NMI} & \text{ACC} & \text{NMI} & \text{ACC} & \text{NMI}\\
    \midrule
    \text{FBOW} & 28.14$\pm$0.89 & 3.26$\pm$0.65 & 12.32$\pm$1.23 & 6.37$\pm$1.53 & 13.92$\pm$1.68 & 8.53$\pm$1.81\\
    \text{FTF-IDF} & 30.58$\pm$1.35 & 7.48$\pm$1.81 & 42.66$\pm$2.72 & 43.79$\pm$3.57 & 25.87$\pm$0.77 & 23.86$\pm$1.42 \\
    \text{FSBERT} & 65.95$\pm$0.00 & 31.55$\pm$0.00 & 60.55$\pm$0.00 & 51.79$\pm$0.00 & 39.50$\pm$0.00 & 32.63$\pm$0.00 \\
    \text{FSCCL} & \underline{81.17$\pm$0.11} & \underline{54.78$\pm$0.18} & \underline{70.45$\pm$1.84} & \underline{61.87$\pm$0.71} & \underline{42.10$\pm$0.72} & \underline{37.40$\pm$0.18}  \\
    \midrule
    \text{\modelname-STC} & 84.38 $\pm$0.92 & 60.75$\pm$1.60 & 77.90 $\pm$1.12 & 67.81$\pm$0.51 & 45.31$\pm$0.75 & 38.25$\pm$0.34  \\
    \text{\modelname-RSTC} & 84.75$\pm$0.60 & 61.33$\pm$0.95 & 78.58$\pm$0.73 & 68.11 $\pm$0.25 & 45.64$\pm$0.58 & 38.36$\pm$0.20 \\
    \text{\modelname} & \textbf{85.10$\pm$0.25}$^*$ & \textbf{62.45$\pm$0.45}$^*$ & \textbf{79.70$\pm$1.13}$^*$ & \textbf{68.83$\pm$0.28}$^*$ & \textbf{46.67$\pm$0.72}$^*$ & \textbf{39.86$\pm$0.58}$^*$ \\
    \bottomrule
  \end{tabular}
  \caption{Experimental results on three short text datasets when $m=4$, where * denotes a significant improvement with \textit{p}-value $<$ 0.01 using the Mann-Whitney U test.
  For all the experiments, we repeat them five times. 
  We bold the \textbf{best result} and underline the \underline{runner-up}.}
  \vspace{-0.3cm}
\label{ta:result}
\end{table*}

\paragraph{Results and discussion.}
The comparison results on three datasets are shown in Table \ref{ta:result}.
From them, we can find that:
(1) Only adopting k-means based federated clustering with traditional text representations (\textbf{FBOW} and \textbf{FTF-IDF}) cannot obtain satisfying results due to the data sparsity problem.
(2) \textbf{FSBERT} outperforms the traditional text representation methods, indicating that adopting pre-trained word embeddings alleviates the sparsity problem, but the fixed SBERT without representation learning cannot obtain discriminative representations for clustering.
(3) \textbf{FSCCL} obtains better clustering results by introducing instance-wise contrastive learning and utilizing the clustering objective proposed in \cite{xie2016unsupervised} for simultaneously representation learning and clustering.
Although \textbf{FSCCL} obtains discriminative representations by deep representation learning, it cannot learn sufficiently discriminative representations due to lacking supervision information, causing limited clustering performance.
(4) \modelname~consistently achieves the best performance, which proves that the robust short text clustering module with generated pseudo-labels as supervision and the federated cluster center aggregation module with efficient communications can significantly improve the federated clustering performance.

\begin{figure*}[t]
  \centering
    \subfigure[FTF-IDF]{
    \begin{minipage}[t]{0.23\linewidth} 
    \includegraphics[width=4.5cm]{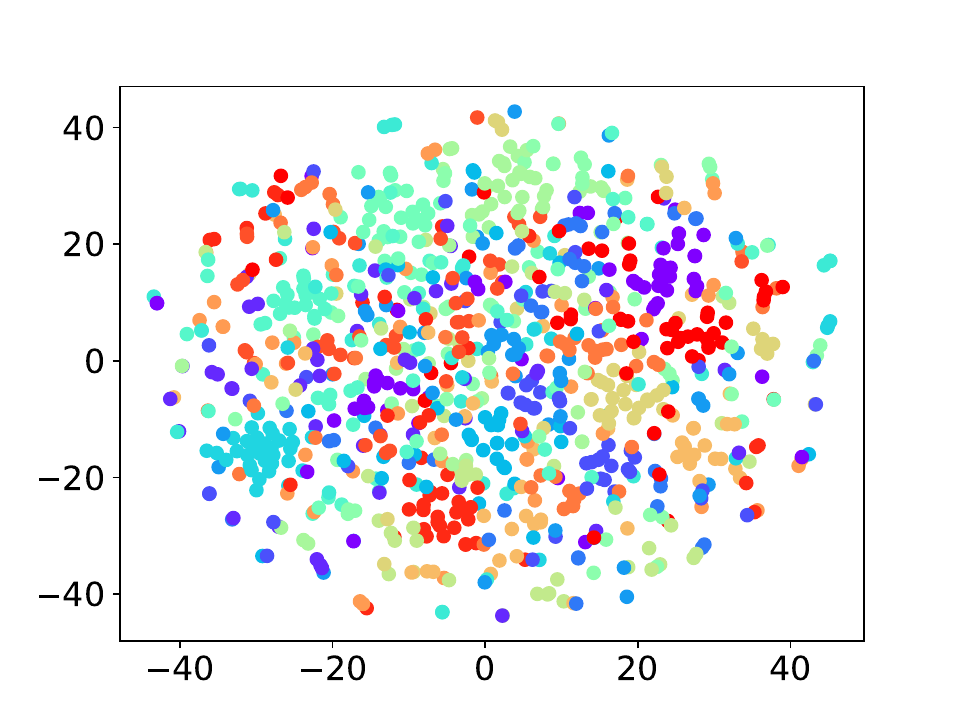}
    \end{minipage}
    }  
    \subfigure[FSBERT]{
    \begin{minipage}[t]{0.23\linewidth} 
    \includegraphics[width=4.5cm]{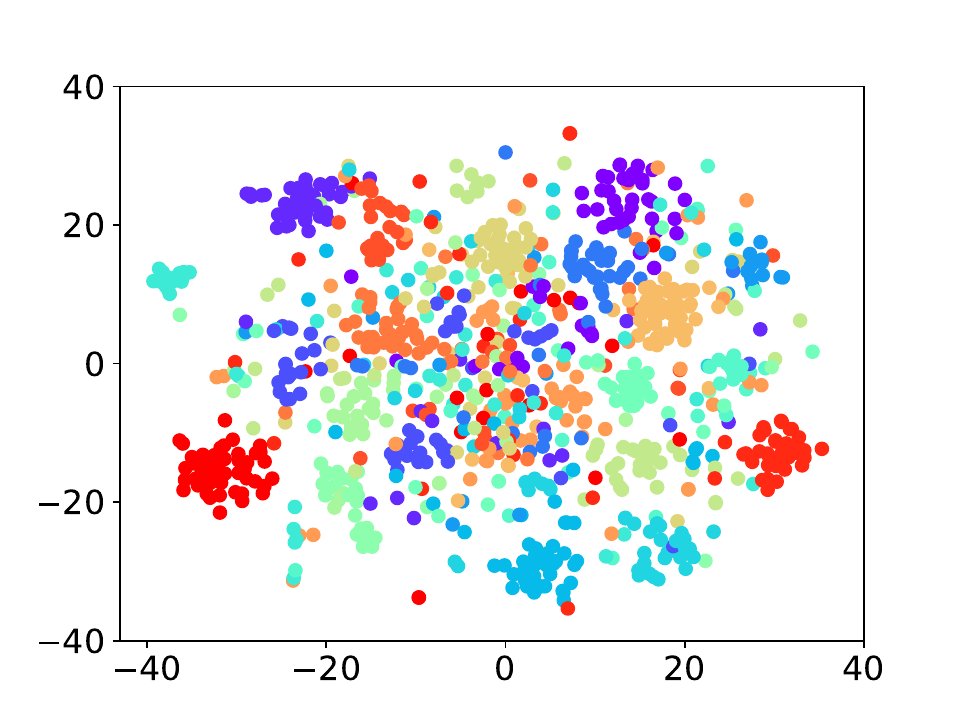}
    \end{minipage}
    }    
    \subfigure[FSCCL]{
    \begin{minipage}[t]{0.23\linewidth} 
    \includegraphics[width=4.5cm]{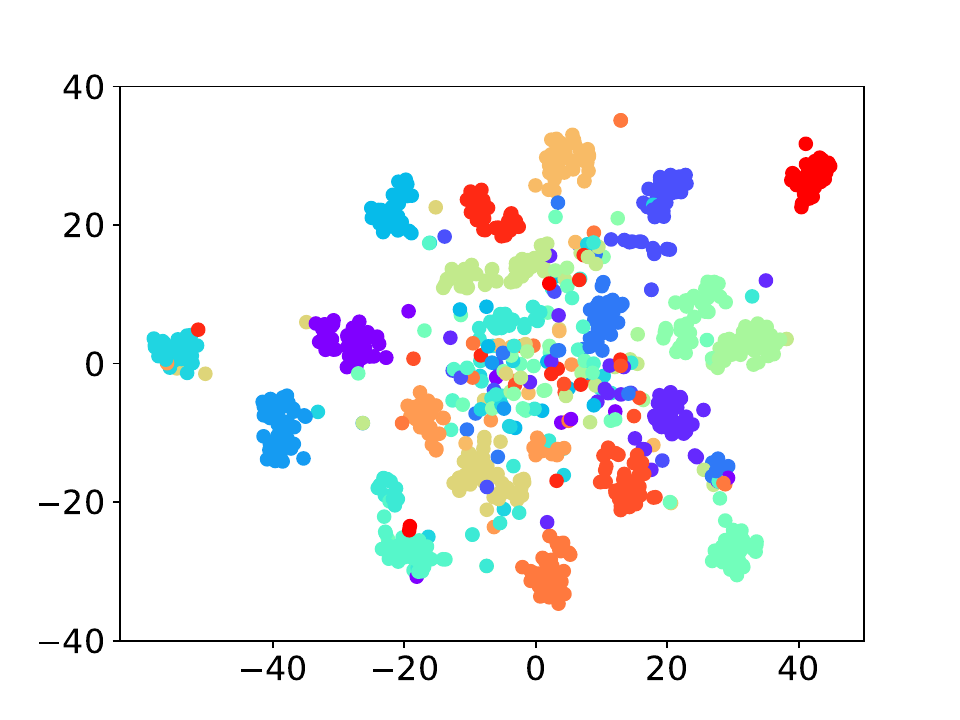}
    \end{minipage}
    }
    \subfigure[\modelname]{
    \begin{minipage}[t]{0.23\linewidth} 
    \includegraphics[width=4.5cm]{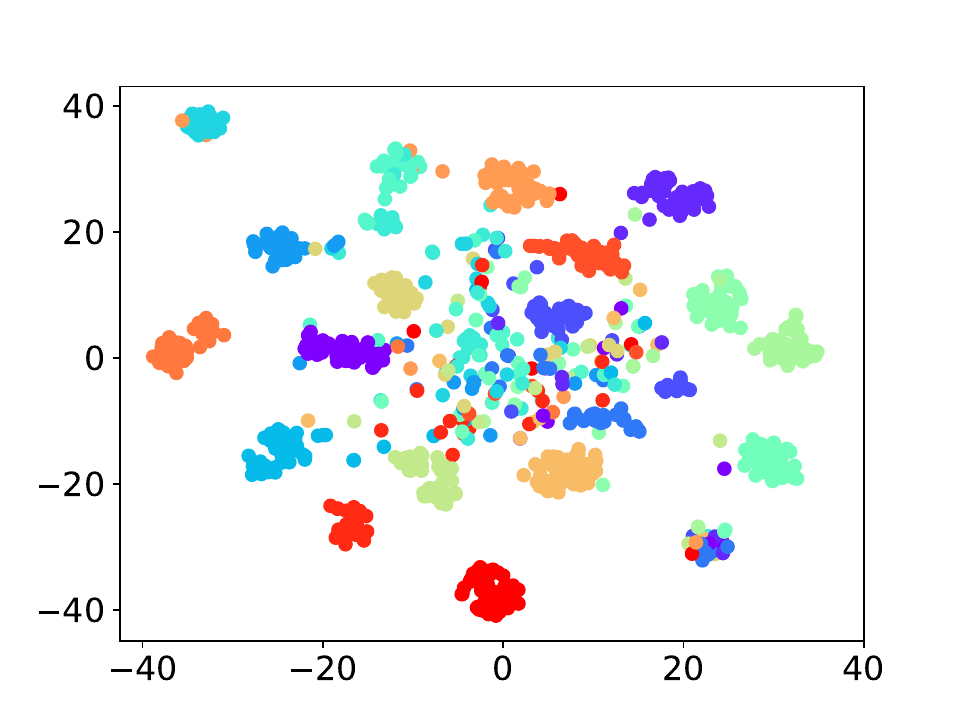}
    \end{minipage}
    }
    \vspace{-0.2cm}
  \caption{TSNE visualization of the representations on Stackoverflow, each color indicates a ground truth category.}
  \vspace{-0.4cm}
  \label{fig:tsne}
\end{figure*}

\paragraph{Visualization.}
To better show the discriminability of text representations, we visualize the representations using t-SNE \cite{van2008visualizing} for \textbf{FTF-IDF}, \textbf{FSBERT}, \textbf{FSCCL}, and \modelname.
The results on \textbf{Stackoverflow} are shown in Fig.\ref{fig:tsne}(a)-(d).
From them, we can see that:
(1) \textbf{FTF-IDF} has no boundaries between clusters, and the points from different clusters have significant overlap.
(2) Although there is less overlap in \textbf{FSBERT}, it still has no significant boundaries between clusters.
(3) \textbf{FSCCL} achieves clear boundaries to some extent, but there are a large proportion of points are grouped to the wrong clusters.
(4) With reliable pseudo-supervised data, \modelname~obtains best text representations with smaller intra-cluster distance, larger inter-cluster distance while more points are grouped to the correct clusters.
The visualization results illustrate the validity of our \modelname~framework.

\subsection{In-depth Analysis (RQ2-RQ5)}
\paragraph{Ablation (RQ2).}
To study how does each component of \modelname~contribute on the final performance, we compare \modelname~with its two variants, including \modelname-STC and \modelname-RSTC.
\modelname-STC only adopts the pseudo-label generation while \modelname-RSTC adopts the pseudo-label generation and the Gaussian-uniform mixture model (i.e., the robust short text clustering module).
The final global model is obtained by averaging the final local models.
The comparison results are shown in Table \ref{ta:result}.
From it, we can observe that
(1) \modelname-STC always get more accurate output predictions than baselines, which indicates that using generated pseudo-labels as supervision information to guide the training is essential.
(2) \modelname-RSTC always get better results than \modelname-STC, indicating that the Gaussian-uniform mixture model is beneficial to obtain more reliable pseudo-supervised data for training.
(2) However, \modelname-RSTC still cannot achieve the best results against \modelname.
Simply averaging the final local models as the global model will sometimes cannot fully exploit all data for local training.
Overall, the above ablation study demonstrates that our proposed robust short text clustering module and federated cluster center aggregation module are effective in solving the FSTC problem.

\begin{figure}[t]
  \centering
    \subfigure[Effects on ACC]{
    \begin{minipage}[t]{0.47\linewidth} 
    \includegraphics[width=4.5cm]{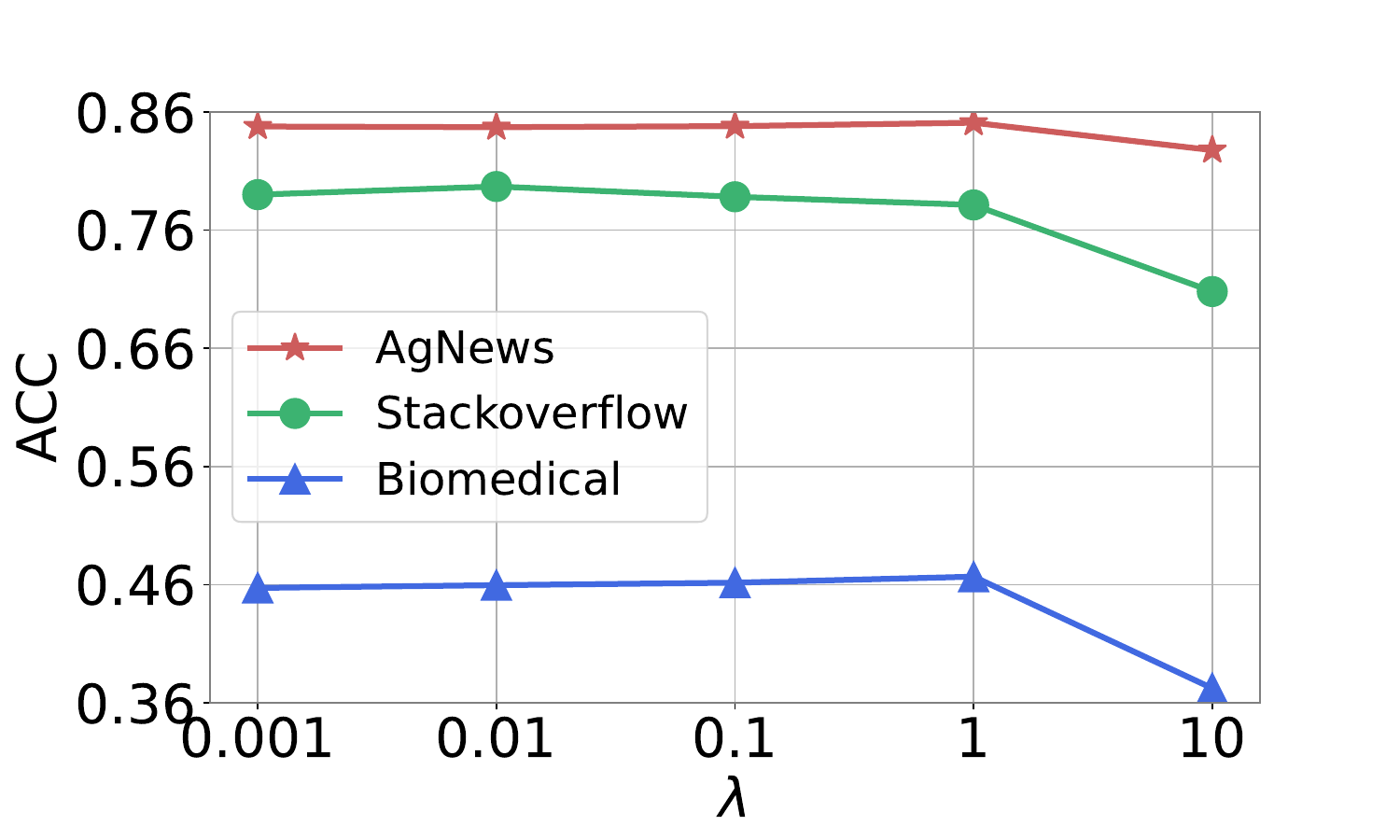}
    \end{minipage}
    }
    \subfigure[Effects on NMI]{
    \begin{minipage}[t]{0.47\linewidth} 
    \includegraphics[width=4.5cm]{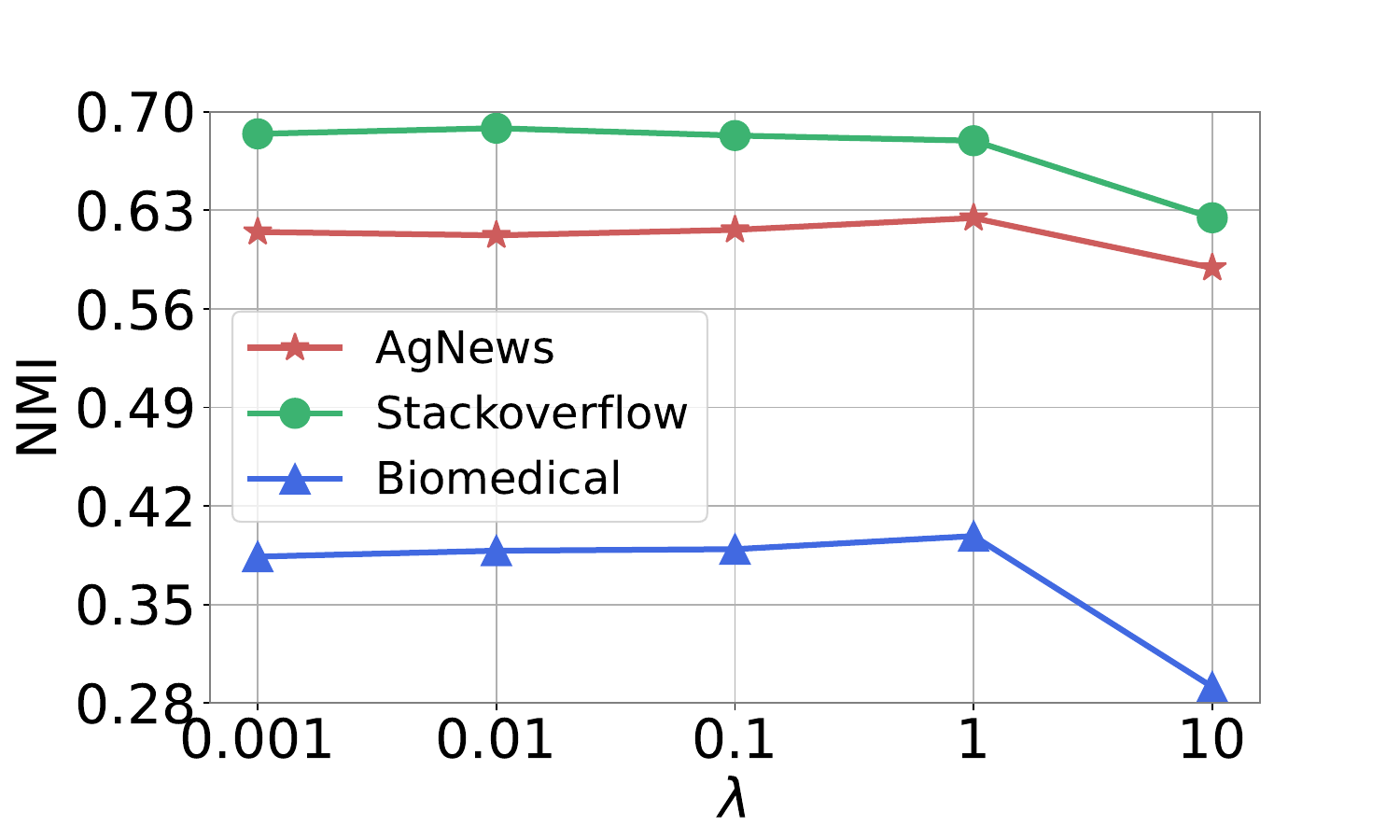}
    \end{minipage}
    }
    \vspace{-0.2cm}
  \caption{The effects of $\lambda$ on model performance when $m=4$.}
  \vspace{-0.2cm}
  \label{fig:para}
\end{figure}

\paragraph{Effect of hyper-parameters (RQ3).}
We first study the effect of $\lambda$ on model performance, where $\lambda$ is a hyper-parameter to balance the clustering loss and the alignment loss in each client.
We vary $\lambda$ in $\{0.001, 0.01, 0.1, 1, 10\}$ and report the results in Fig.\ref{fig:para}.
Fig.\ref{fig:para} shows that the performance first gradually increases and then decreases.
It indicates that when $\lambda$ approaches $0$, the alignment loss cannot produce sufficiently positive effects. When $\lambda$ becomes too large, the alignment loss will suppress the clustering loss, which also reduces the clustering performance.
Empirically, we choose $\lambda=0.01$ on \textbf{Stackoverflow} while $\lambda=1$ on \textbf{AgNews} and \textbf{Biomedical}.

\begin{figure}[t]
  \centering
    \subfigure[Effects of $m$]{
    \begin{minipage}[t]{0.47\linewidth} 
    \includegraphics[width=4.5cm]{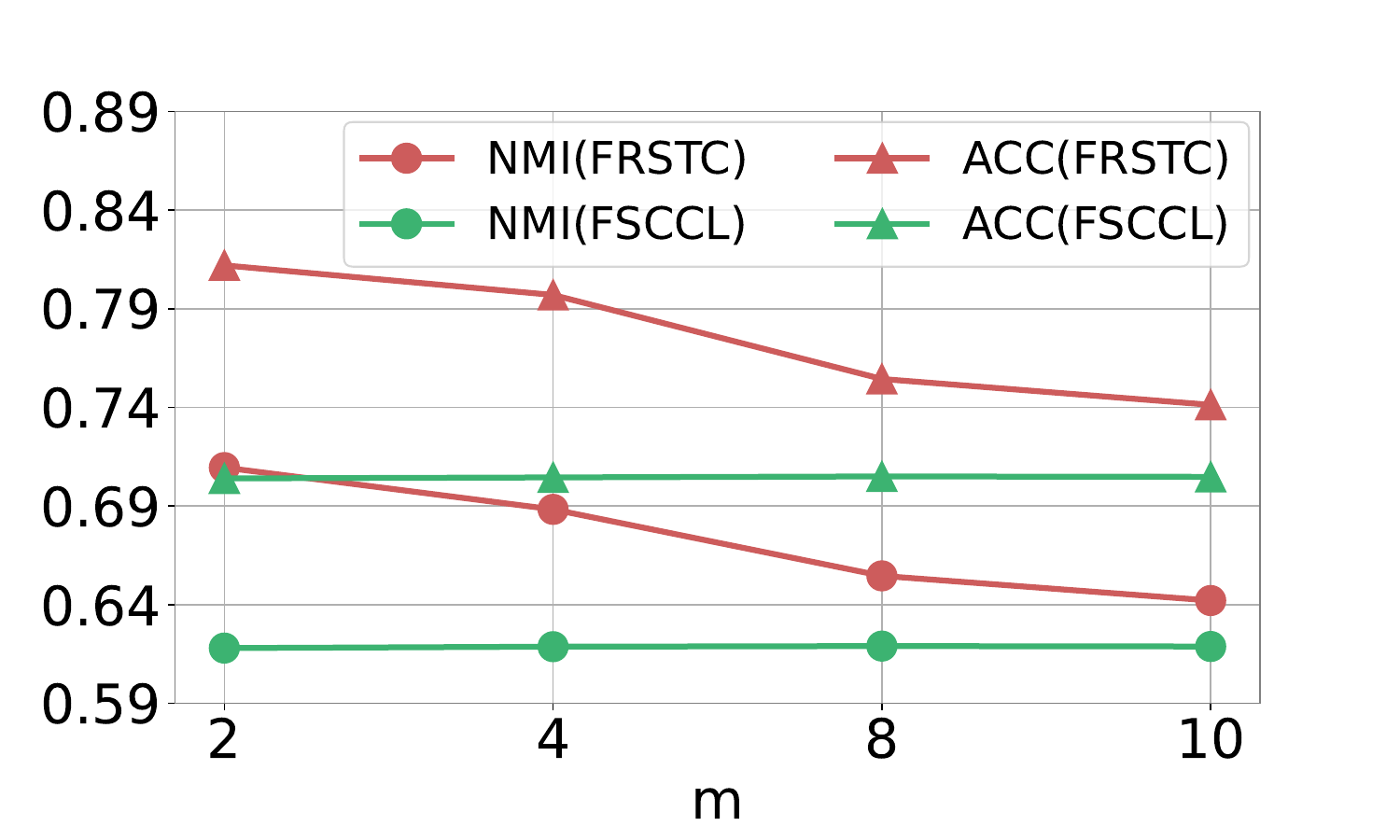}
    \end{minipage}
    }
    \subfigure[Effects of $\rho$]{
    \begin{minipage}[t]{0.47\linewidth} 
    \includegraphics[width=4.5cm]{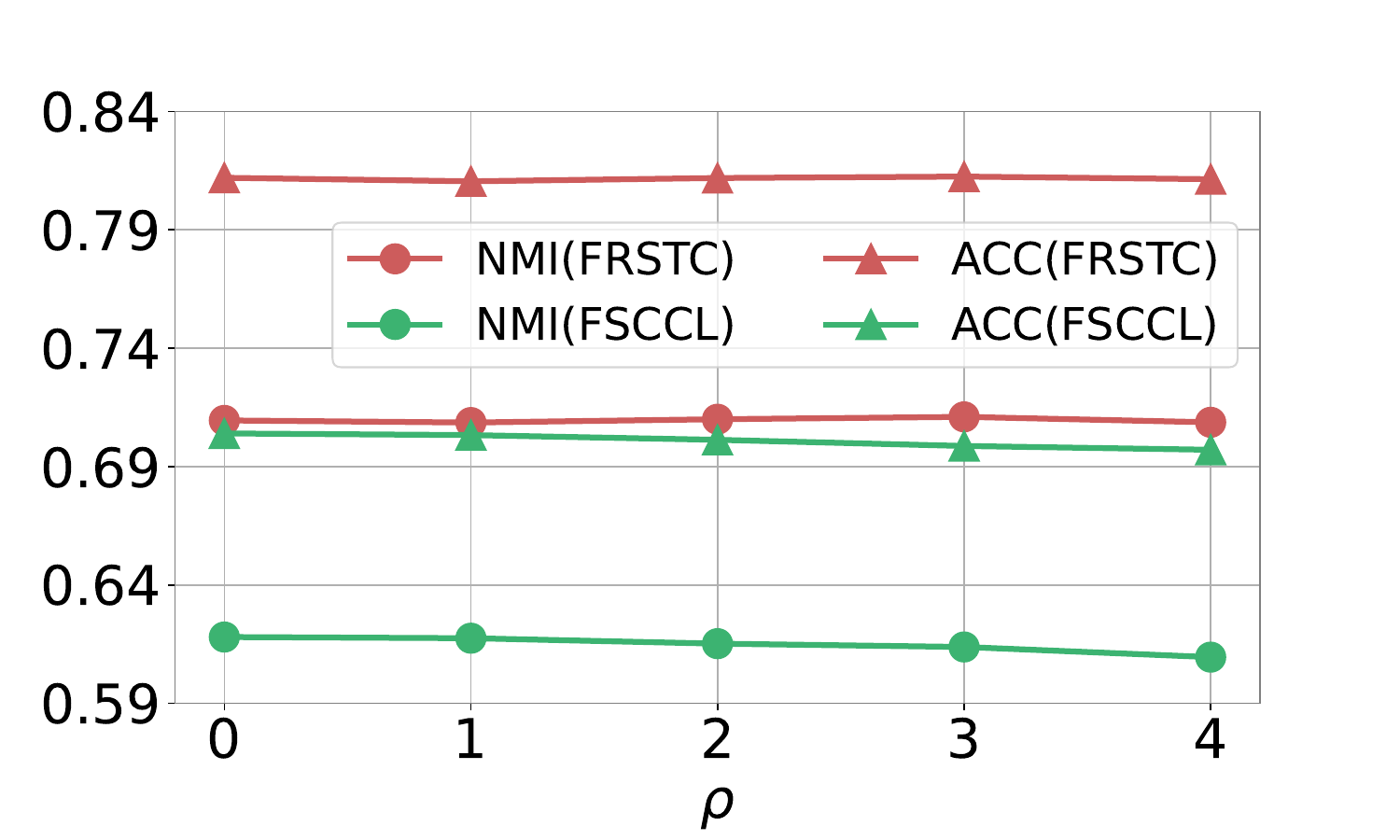}
    \end{minipage}
    }
    \vspace{-0.2cm}
  \caption{The effects of $m$ and $\rho$ on model performance.}
  \vspace{-0.2cm}
  \label{fig:clientsandnoniid}
\end{figure}

\paragraph{Effect of client number (RQ4).}
We also study the effect of client number $m$ on \modelname~and \textbf{FSCCL} by varying $m$ in $\{2, 4, 8, 10\}$ and report the results in Fig.\ref{fig:clientsandnoniid}(a) on \textbf{Stackoverflow}.
Fig.\ref{fig:clientsandnoniid}(a) shows that \modelname~keeps better performance than \textbf{FSCCL} all the time, which illustrates the validity of our framework.
Besides, it is a normal phenomenon that the performance of \modelname~declines as $m$ increases, since the number of samples in each client decreases.
The reason why the performance of \textbf{FSCCL} remains relatively stable as $m$ increases may be that SCCL cannot obtain better performance with more samples in a client.

\paragraph{Effect of non-IID level (RQ5).}
We finally study the effect of non-IID level $\rho$ on \modelname~and \textbf{FSCCL} by varying $\rho$ in $\{0, 1, 2, 3, 4\}$ and report the results in Fig.\ref{fig:clientsandnoniid}(b) on \textbf{Stackoverflow}, where $\rho=0$ denotes IID.
Fig.\ref{fig:clientsandnoniid} shows that \modelname~keeps better performance than \textbf{FSCCL} all the time, which illustrates the effectiveness of our framework.
Besides, the performance of both models remains relatively stable as $\rho$ increases, which indicates that our framework is robust to different non-IID levels with more efficient communications.

\section{Conclusion}
In this paper, we propose a federated robust short text clustering framework (\modelname). which includes the robust short text clustering module and the federated cluster center aggregation module.
To our best knowledge, we are the first to address short text clustering problem in the federated setting.
Moreover, we innovatively combine optimal transport to generate pseudo-labels with Gaussian-uniform mixture model to improve the reliability of the pseudo-supervised data.
We also conduct extensive experiments to demonstrate the superior performance of our proposed \modelname~on several real-world datasets.

\bibliographystyle{named}
\bibliography{ijcai23}

\begin{thebibliography}{}

\bibitem[\protect\citeauthoryear{Asano \bgroup \em et al.\egroup
  }{2020}]{asano2019self}
Yuki~Markus Asano, Christian Rupprecht, and Andrea Vedaldi.
\newblock Self-labelling via simultaneous clustering and representation
  learning.
\newblock In {\em 8th International Conference on Learning Representations,
  {ICLR} 2020, Addis Ababa, Ethiopia, April 26-30, 2020}. OpenReview.net, 2020.

\bibitem[\protect\citeauthoryear{Bridle}{1990}]{bridle1990probabilistic}
John~S Bridle.
\newblock Probabilistic interpretation of feedforward classification network
  outputs, with relationships to statistical pattern recognition.
\newblock In {\em Neurocomputing}, pages 227--236. Springer, 1990.

\bibitem[\protect\citeauthoryear{Caron \bgroup \em et al.\egroup
  }{2020}]{caron2020unsupervised}
Mathilde Caron, Ishan Misra, Julien Mairal, Priya Goyal, Piotr Bojanowski, and
  Armand Joulin.
\newblock Unsupervised learning of visual features by contrasting cluster
  assignments.
\newblock {\em Advances in Neural Information Processing Systems},
  33:9912--9924, 2020.

\bibitem[\protect\citeauthoryear{Chung \bgroup \em et al.\egroup
  }{2022}]{chung2022federated}
Jichan Chung, Kangwook Lee, and Kannan Ramchandran.
\newblock Federated unsupervised clustering with generative models.
\newblock In {\em AAAI 2022 International Workshop on Trustable, Verifiable and
  Auditable Federated Learning}, 2022.

\bibitem[\protect\citeauthoryear{Coretto and Hennig}{2016}]{coretto2016robust}
Pietro Coretto and Christian Hennig.
\newblock Robust improper maximum likelihood: tuning, computation, and a
  comparison with other methods for robust gaussian clustering.
\newblock {\em Journal of the American Statistical Association},
  111(516):1648--1659, 2016.

\bibitem[\protect\citeauthoryear{Cuturi}{2013}]{cuturi2013sinkhorn}
Marco Cuturi.
\newblock Sinkhorn distances: Lightspeed computation of optimal transport.
\newblock {\em Advances in neural information processing systems}, 26, 2013.

\bibitem[\protect\citeauthoryear{Dennis \bgroup \em et al.\egroup
  }{2021}]{dennis2021heterogeneity}
Don~Kurian Dennis, Tian Li, and Virginia Smith.
\newblock Heterogeneity for the win: One-shot federated clustering.
\newblock In {\em International Conference on Machine Learning}, pages
  2611--2620. PMLR, 2021.

\bibitem[\protect\citeauthoryear{Ghosh \bgroup \em et al.\egroup
  }{2017}]{ghosh2017robust}
Aritra Ghosh, Himanshu Kumar, and P.~S. Sastry.
\newblock Robust loss functions under label noise for deep neural networks.
\newblock In Satinder Singh and Shaul Markovitch, editors, {\em Proceedings of
  the Thirty-First {AAAI} Conference on Artificial Intelligence, February 4-9,
  2017, San Francisco, California, {USA}}, pages 1919--1925. {AAAI} Press,
  2017.

\bibitem[\protect\citeauthoryear{Ghosh \bgroup \em et al.\egroup
  }{2020}]{ghosh2020efficient}
Avishek Ghosh, Jichan Chung, Dong Yin, and Kannan Ramchandran.
\newblock An efficient framework for clustered federated learning.
\newblock {\em Advances in Neural Information Processing Systems},
  33:19586--19597, 2020.

\bibitem[\protect\citeauthoryear{Gu \bgroup \em et al.\egroup
  }{2020}]{gu2020spherical}
Xiang Gu, Jian Sun, and Zongben Xu.
\newblock Spherical space domain adaptation with robust pseudo-label loss.
\newblock In {\em Proceedings of the IEEE/CVF Conference on Computer Vision and
  Pattern Recognition}, pages 9101--9110, 2020.

\bibitem[\protect\citeauthoryear{Hadifar \bgroup \em et al.\egroup
  }{2019}]{hadifar2019self}
Amir Hadifar, Lucas Sterckx, Thomas Demeester, and Chris Develder.
\newblock A self-training approach for short text clustering.
\newblock In {\em Proceedings of the 4th Workshop on Representation Learning
  for NLP (RepL4NLP-2019)}, pages 194--199, 2019.

\bibitem[\protect\citeauthoryear{Hu \bgroup \em et al.\egroup
  }{2021}]{hu2021learning}
Weibo Hu, Chuan Chen, Fanghua Ye, Zibin Zheng, and Yunfei Du.
\newblock Learning deep discriminative representations with pseudo supervision
  for image clustering.
\newblock {\em Information Sciences}, 568:199--215, 2021.

\bibitem[\protect\citeauthoryear{Kingma and Ba}{2015}]{kingma2014adam}
Diederik~P. Kingma and Jimmy Ba.
\newblock Adam: {A} method for stochastic optimization.
\newblock In Yoshua Bengio and Yann LeCun, editors, {\em 3rd International
  Conference on Learning Representations, {ICLR} 2015, San Diego, CA, USA, May
  7-9, 2015, Conference Track Proceedings}, 2015.

\bibitem[\protect\citeauthoryear{Kumar \bgroup \em et al.\egroup
  }{2020}]{kumar2020federated}
Hemant~H Kumar, VR~Karthik, and Mydhili~K Nair.
\newblock Federated k-means clustering: A novel edge ai based approach for
  privacy preservation.
\newblock In {\em 2020 IEEE International Conference on Cloud Computing in
  Emerging Markets (CCEM)}, pages 52--56. IEEE, 2020.

\bibitem[\protect\citeauthoryear{Lathuili{\`{e}}re \bgroup \em et al.\egroup
  }{2018}]{deepgum}
St{\'{e}}phane Lathuili{\`{e}}re, Pablo Mesejo, Xavier Alameda{-}Pineda, and
  Radu Horaud.
\newblock Deepgum: Learning deep robust regression with a gaussian-uniform
  mixture model.
\newblock In Vittorio Ferrari, Martial Hebert, Cristian Sminchisescu, and Yair
  Weiss, editors, {\em Computer Vision - {ECCV} 2018 - 15th European
  Conference, Munich, Germany, September 8-14, 2018, Proceedings, Part {V}},
  volume 11209 of {\em Lecture Notes in Computer Science}, pages 205--221.
  Springer, 2018.

\bibitem[\protect\citeauthoryear{Li \bgroup \em et al.\egroup
  }{2022}]{li2022unsupervised}
Jinning Li, Huajie Shao, Dachun Sun, Ruijie Wang, Yuchen Yan, Jinyang Li,
  Shengzhong Liu, Hanghang Tong, and Tarek Abdelzaher.
\newblock Unsupervised belief representation learning with
  information-theoretic variational graph auto-encoders.
\newblock In {\em Proceedings of the 45th International ACM SIGIR Conference on
  Research and Development in Information Retrieval}, pages 1728--1738, 2022.

\bibitem[\protect\citeauthoryear{Magdziarczyk}{2019}]{magdziarczyk2019right}
Malgorzata Magdziarczyk.
\newblock Right to be forgotten in light of regulation (eu) 2016/679 of the
  european parliament and of the council of 27 april 2016 on the protection of
  natural persons with regard to the processing of personal data and on the
  free movement of such data, and repealing directive 95/46/ec.
\newblock In {\em 6th INTERNATIONAL MULTIDISCIPLINARY SCIENTIFIC CONFERENCE ON
  SOCIAL SCIENCES AND ART SGEM 2019}, pages 177--184, 2019.

\bibitem[\protect\citeauthoryear{McMahan \bgroup \em et al.\egroup
  }{2017a}]{DBLP:conf/aistats/McMahanMRHA17}
Brendan McMahan, Eider Moore, Daniel Ramage, Seth Hampson, and
  Blaise~Ag{\"{u}}era y~Arcas.
\newblock Communication-efficient learning of deep networks from decentralized
  data.
\newblock In Aarti Singh and Xiaojin~(Jerry) Zhu, editors, {\em Proceedings of
  the 20th International Conference on Artificial Intelligence and Statistics,
  {AISTATS} 2017, 20-22 April 2017, Fort Lauderdale, FL, {USA}}, volume~54 of
  {\em Proceedings of Machine Learning Research}, pages 1273--1282. {PMLR},
  2017.

\bibitem[\protect\citeauthoryear{McMahan \bgroup \em et al.\egroup
  }{2017b}]{mcmahan2017communication}
Brendan McMahan, Eider Moore, Daniel Ramage, Seth Hampson, and Blaise~Aguera
  y~Arcas.
\newblock Communication-efficient learning of deep networks from decentralized
  data.
\newblock In {\em Artificial intelligence and statistics}, pages 1273--1282.
  PMLR, 2017.

\bibitem[\protect\citeauthoryear{Mikolov \bgroup \em et al.\egroup
  }{2013}]{mikolov2013distributed}
Tomas Mikolov, Ilya Sutskever, Kai Chen, Greg~S Corrado, and Jeff Dean.
\newblock Distributed representations of words and phrases and their
  compositionality.
\newblock {\em Advances in neural information processing systems}, 26, 2013.

\bibitem[\protect\citeauthoryear{Otto}{2018}]{otto2018regulation}
Marta Otto.
\newblock Regulation (eu) 2016/679 on the protection of natural persons with
  regard to the processing of personal data and on the free movement of such
  data (general data protection regulation--gdpr).
\newblock In {\em International and European Labour Law}, pages 958--981. Nomos
  Verlagsgesellschaft mbH \& Co. KG, 2018.

\bibitem[\protect\citeauthoryear{Papadimitriou and
  Steiglitz}{1998}]{papadimitriou1998combinatorial}
Christos~H Papadimitriou and Kenneth Steiglitz.
\newblock {\em Combinatorial optimization: algorithms and complexity}.
\newblock Courier Corporation, 1998.

\bibitem[\protect\citeauthoryear{Paszke \bgroup \em et al.\egroup
  }{2019}]{pytorch}
Adam Paszke, Sam Gross, Francisco Massa, Adam Lerer, James Bradbury, Gregory
  Chanan, Trevor Killeen, Zeming Lin, Natalia Gimelshein, Luca Antiga, Alban
  Desmaison, Andreas Kopf, Edward Yang, Zachary DeVito, Martin Raison, Alykhan
  Tejani, Sasank Chilamkurthy, Benoit Steiner, Lu~Fang, Junjie Bai, and Soumith
  Chintala.
\newblock Pytorch: An imperative style, high-performance deep learning library.
\newblock In {\em Advances in Neural Information Processing Systems 32}, pages
  8024--8035. Curran Associates, Inc., 2019.

\bibitem[\protect\citeauthoryear{Pedrycz}{2021}]{pedrycz2021federated}
Witold Pedrycz.
\newblock Federated fcm: Clustering under privacy requirements.
\newblock {\em IEEE Transactions on Fuzzy Systems}, 2021.

\bibitem[\protect\citeauthoryear{Rakib \bgroup \em et al.\egroup
  }{2020}]{rakib2020enhancement}
Md~Rashadul~Hasan Rakib, Norbert Zeh, Magdalena Jankowska, and Evangelos
  Milios.
\newblock Enhancement of short text clustering by iterative classification.
\newblock In {\em International Conference on Applications of Natural Language
  to Information Systems}, pages 105--117. Springer, 2020.

\bibitem[\protect\citeauthoryear{Reimers and
  Gurevych}{2019}]{reimers2019sentence}
Nils Reimers and Iryna Gurevych.
\newblock Sentence-bert: Sentence embeddings using siamese bert-networks.
\newblock In {\em Proceedings of the 2019 Conference on Empirical Methods in
  Natural Language Processing and the 9th International Joint Conference on
  Natural Language Processing (EMNLP-IJCNLP)}, pages 3982--3992, 2019.

\bibitem[\protect\citeauthoryear{Salton and
  McGill}{1983}]{salton1983introduction}
Gerard Salton and Michael~J McGill.
\newblock {\em Introduction to modern information retrieval}.
\newblock mcgraw-hill, 1983.

\bibitem[\protect\citeauthoryear{Scott and Matwin}{1998}]{scott1998text}
Sam Scott and Stan Matwin.
\newblock Text classification using wordnet hypernyms.
\newblock In {\em Usage of WordNet in natural language processing systems},
  1998.

\bibitem[\protect\citeauthoryear{Stallmann and Wilbik}{2022}]{ffcm2022morris}
Morris Stallmann and Anna Wilbik.
\newblock Towards federated clustering: {A} federated fuzzy c-means algorithm
  {(FFCM)}.
\newblock {\em CoRR}, abs/2201.07316, 2022.

\bibitem[\protect\citeauthoryear{Stieglitz \bgroup \em et al.\egroup
  }{2018}]{stieglitz2018social}
Stefan Stieglitz, Milad Mirbabaie, Bj{\"o}rn Ross, and Christoph Neuberger.
\newblock Social media analytics--challenges in topic discovery, data
  collection, and data preparation.
\newblock {\em International journal of information management}, 39:156--168,
  2018.

\bibitem[\protect\citeauthoryear{Tan \bgroup \em et al.\egroup
  }{2022}]{tan2022fedproto}
Yue Tan, Guodong Long, Lu~Liu, Tianyi Zhou, Qinghua Lu, Jing Jiang, and Chengqi
  Zhang.
\newblock Fedproto: Federated prototype learning across heterogeneous clients.
\newblock In {\em AAAI Conference on Artificial Intelligence}, volume~1,
  page~3, 2022.

\bibitem[\protect\citeauthoryear{Van~der Maaten and
  Hinton}{2008}]{van2008visualizing}
Laurens Van~der Maaten and Geoffrey Hinton.
\newblock Visualizing data using t-sne.
\newblock {\em Journal of machine learning research}, 9(11), 2008.

\bibitem[\protect\citeauthoryear{Wu \bgroup \em et al.\egroup
  }{2022}]{wu2022personalized}
Chuhan Wu, Fangzhao Wu, Yongfeng Huang, and Xing Xie.
\newblock Personalized news recommendation: Methods and challenges.
\newblock {\em ACM Transactions on Information Systems (TOIS)}, 2022.

\bibitem[\protect\citeauthoryear{Xie \bgroup \em et al.\egroup
  }{2016}]{xie2016unsupervised}
Junyuan Xie, Ross Girshick, and Ali Farhadi.
\newblock Unsupervised deep embedding for clustering analysis.
\newblock In {\em International conference on machine learning}, pages
  478--487. PMLR, 2016.

\bibitem[\protect\citeauthoryear{Xu \bgroup \em et al.\egroup
  }{2017}]{xu2017self}
Jiaming Xu, Bo~Xu, Peng Wang, Suncong Zheng, Guanhua Tian, Jun Zhao, and Bo~Xu.
\newblock Self-taught convolutional neural networks for short text clustering.
\newblock {\em Neural Networks}, 88:22--31, 2017.

\bibitem[\protect\citeauthoryear{Zhang \bgroup \em et al.\egroup
  }{2015}]{zhang2015character}
Xiang Zhang, Junbo Zhao, and Yann LeCun.
\newblock Character-level convolutional networks for text classification.
\newblock {\em Advances in neural information processing systems}, 28, 2015.

\bibitem[\protect\citeauthoryear{Zhang \bgroup \em et al.\egroup
  }{2021}]{zhang2021supporting}
Dejiao Zhang, Feng Nan, Xiaokai Wei, Shang{-}Wen Li, Henghui Zhu, Kathleen~R.
  McKeown, Ramesh Nallapati, Andrew~O. Arnold, and Bing Xiang.
\newblock Supporting clustering with contrastive learning.
\newblock In Kristina Toutanova, Anna Rumshisky, Luke Zettlemoyer, Dilek
  Hakkani{-}T{\"{u}}r, Iz~Beltagy, Steven Bethard, Ryan Cotterell, Tanmoy
  Chakraborty, and Yichao Zhou, editors, {\em Proceedings of the 2021
  Conference of the North American Chapter of the Association for Computational
  Linguistics: Human Language Technologies, {NAACL-HLT} 2021, Online, June
  6-11, 2021}, pages 5419--5430. Association for Computational Linguistics,
  2021.

\end{thebibliography}
\end{document}